\documentclass[conference]{IEEEtran}
\IEEEoverridecommandlockouts
\usepackage{cite}
\usepackage{amsmath,amssymb,amsfonts}

\usepackage{algorithmicx}
\usepackage{algpseudocode}

\usepackage{booktabs}
\usepackage{tabto}
\usepackage[ruled]{algorithm2e}
\usepackage{graphicx}
\usepackage{textcomp}
\usepackage{xcolor}
\def\BibTeX{{\rm B\kern-.05em{\sc i\kern-.025em b}\kern-.08em
    T\kern-.1667em\lower.7ex\hbox{E}\kern-.125emX}}
\begin{document}

\title{Privacy-Preserving Heterogeneous Federated Learning for Sensitive Healthcare Data\\
}


\author{Yukai Xu
\\
\IEEEauthorblockA{
\textit{Kyushu University}
}
\and
Jingfeng Zhang
\\
\IEEEauthorblockA{
\textit{The University of Auckland / RIKEN}
}
\and
Yujie Gu
\\
\IEEEauthorblockA{
\textit{Kyushu University}
}
}

\maketitle

\begin{abstract}
\if
Machine learning has attained remarkable achievements in a wide range of real-world applications. 
Yet, in the realm of healthcare with decentralized facilities, two major challenges arise concerning the protection of data and models.
The data-level challenge concerns the data privacy leakage when centralizing data with sensitive personal information. 
Another model-level challenge arises from the heterogeneity of local models, which need to be collaboratively trained while ensuring their confidentiality to address intellectual property concerns.

In this paper, we propose a new framework termed \textit{Abstention-Aware Federated Voting} (AAFV) that can collaboratively and confidentially train heterogeneous local models while simultaneously protecting the data privacy. 
This is achieved by integrating a novel abstention-aware voting mechanism 
and a differential privacy mechanism onto local models' predictions.
In particular, 
the proposed abstention-aware voting mechanism exploits a threshold-based abstention method to select high-confidence votes from heterogeneous local models, which not only enhances the learning utility but also protects model confidentiality. 
Furthermore, we implement AAFV on two practical prediction tasks of diabetes and in-hospital patient mortality. 
The experiments demonstrate the effectiveness and confidentiality of AAFV in terms of both testing accuracy and privacy strength. 
\fi
In the realm of healthcare where decentralized facilities are prevalent, machine learning faces two major challenges concerning the protection of data and models. 
The data-level challenge concerns the data privacy leakage when centralizing data with sensitive personal information. 
While the model-level challenge arises from the heterogeneity of local models, which need to be collaboratively trained while ensuring their confidentiality to address intellectual property concerns.
To tackle these challenges, we propose a new framework termed \textit{Abstention-Aware Federated Voting} (AAFV) that can collaboratively and confidentially train heterogeneous local models while simultaneously protecting the data privacy.
This is achieved by integrating a novel abstention-aware voting mechanism 
and a differential privacy mechanism onto local models' predictions.
In particular, the proposed abstention-aware voting mechanism exploits a threshold-based abstention method to select high-confidence votes from heterogeneous local models, which not only enhances the learning utility but also protects model confidentiality.
Furthermore, we implement AAFV on two practical prediction tasks of diabetes and in-hospital patient mortality. 
The experiments demonstrate the effectiveness and confidentiality of AAFV in testing accuracy and privacy protection. 

\if
When training machine learning models with healthcare data, significant challenges arise, particularly in terms of privacy concerns with merging sensitive data from multiple sources, and intellectual property issues when clients are required to share their models.
Federated learning offers a solution to privacy concerns by combining models trained on distributed healthcare data without the need to share the actual data. 
However, traditional federated learning approaches, which average parameters from models trained by various clients in a ``white box'' setting, don't fully address intellectual property worries or account for different model types.

In this research, we propose a new federated learning framework that enables the collaborative training of heterogeneous local models without the need to disclose their specific parameters. 
This ``black box'' approach preserves the privacy of local model details and  effectively addresses concerns around intellectual property. 
Moreover, we run practical experiments demonstrating that our framework outperforms methods training models in isolation.
\fi
\end{abstract}

\section{Introduction}
Machine learning nowadays has achieved tremendous success in real-world applications, such as 
medical image diagnose \cite{cnnmedicalimage,dlimage,transferimage}, disease risk prediction \cite{superpred,pregnant,heartpred}, and genetic information analyse \cite{persongene,genorisk,dataming}, etc.

In the healthcare industry, an extensive dataset providing a comprehensive multi-faceted perspective is typically required to guarantee model utility. However, healthcare data collected by medical and health institutions across different regions often have inherent biases. For example, data on cystic fibrosis in the US and UK might differ due to varying treatment methods \cite{cf}. Similarly, institutions specializing in oncology and dentistry might record data on diabetic patients with certain biases in various aspects \cite{diabeteor}. These biases could tremendously reduce the model utility by leading to skewed or unrepresentative results.

This makes it necessary to integrate more institutions and to leverage a broader range of data, thereby increasing the dataset size and mitigating inherent biases in the data sources.
However, collecting the healthcare data faces legal, social, and privacy challenges. 
Regulations such as the US Health Insurance Portability and Accountability Act (HIPAA) and the EU General Data Protection Regulation (GDPR) \cite{gdpr}, which strictly control the use of personal health information, pose challenges to data sharing and access in healthcare.

Federated learning enables a group of healthcare facilities to collaboratively train a machine learning model using data from various sources, without the need to share sensitive information. This approach has been applied in tasks like predicting Adverse Drug Reactions and differentiating brain cancer MRI images \cite{Fedavgreaction,fedavgMedicine}. Currently, applications of federated learning in healthcare often utilize an algorithm known as Federated Averaging (FedAvg) \cite{FedAvg}. Within this framework, a central server distributes a global model to each facility (client), and aggregates locally trained model parameters to form a final averaged global model. However, FedAvg struggles with model heterogeneity, as all models used are distributed by the central server. Furthermore, some confidential models cannot be freely distributed to unauthorized facilities, further restricting the usage of FedAvg.

In many practical collaborative learning scenarios, local clients/hospitals typically develop their local models using private resources (e.g. patient medical records) and proprietary techniques. This leads to heterogeneity among local models and gives rise to challenges related to data privacy and model intellectual property concerns. Even though federated learning offers a distributed scheme where local models can be trained without sharing their private datasets, it has been demonstrated that certain attacks, such as membership inference attacks \cite{memberattack,efficientinferattack}, can infer training data from information like predictions, gradients, and model parameters.

Differential privacy provides a rigorous mathematical framework for protecting data privacy in machine learning, see e.g. \cite{dpepsilon,dp,dpsgd}. Several methods, including the exponential mechanism \cite{expodp}, objective perturbation \cite{dperm, dpregression}, and the piecewise mechanism \cite{pmdp}, have been proposed to meet differential privacy requirements. In the healthcare realm, differential privacy mechanisms have been implemented to protect sensitive healthcare datasets against potential attacks in settings where all clients have the same structure \cite{dphealth, dpfdhealth}. 
However, 
it is still quite challenging to simultaneously deal with the model heterogeneity and data privacy issues in the practical healthcare applications.

In this paper, to address these challenges, we propose a novel framework, termed Abstention-Aware Federated Voting (AAFV). This framework enables collaborative and confidential training of heterogeneous local models while protecting data privacy. This is achieved by integrating a novel abstention-aware voting mechanism with a local differential privacy mechanism into collaborative learning.  In particular, the proposed abstention-aware voting mechanism utilizes a threshold-based abstention method to select high-confidence votes from heterogeneous local models, which not only enhances learning utility but also protects model confidentiality.

In summary, the contributions of this paper are as below:
\begin{itemize}
    \item We introduce a new Abstention-Aware Federated Voting (AAFV) framework, designed to enable distributed healthcare facilities to collaboratively and confidentially train heterogeneous local models. 
    This framework not only preserves sensitive patient data privacy but also protects the model intellectual property.
    \item 
    Empirical experiments are conducted to evaluate the performance of the AAFV framework in two critical real-world healthcare tasks: diabetes prediction and in-hospital patient mortality prediction. 
    The results demonstrate the effectiveness and superiority of AAFV in practical applications, particularly in enhancing model utility while protecting data privacy and model confidentiality. 
\end{itemize}

\section{Preliminaries}
\label{sec:preliminary}

\subsection{Problem Formulation}
Typically, healthcare risk assessment  can be formulated as a binary classification task: either a patient is positive or negative to a certain risk. 
The $i$-th sample $\mathbf{x}_i$ in a healthcare dataset is represented in terms of $C$ features: $\mathbf{x}_i \in \mathbb{R}^C$. The label of $\mathbf{x}_i$ is denoted as $y_i \in \{0, 1\}$, where $1$ and $0$ represent the positive and negative labels, respectively. 

Consider a federated learning system with $K$ distributed local health entities/clients. 
The $k$-th client possesses a private labeled dataset $\mathcal{D}_k =\{(\mathbf{x}_i, y_i)\}_{i=1}^{N_k}$, which contains $N_k$ samples as well as the corresponding diagnosis/labels. 
There is a public unlabeled dataset $\mathcal{D}^u = \{\mathbf{x}_i^u\}_{i=1}^{N_u}$ that includes only $N_u$ samples of features without labels. 
The public dataset will be exploited for aggregation and is only accessible for all entities participated in the federated learning system. 

The $k$-th client owns a local machine learning model $f_k: \mathbb{R}^C \to [0,1]$ for healthcare data analysis, which takes a healthcare data sample $\mathbf{x}\in\mathbb{R}^C$ as the input and outputs a real number $p_k=f_k(\mathbf{x}) \in [0,1]$ representing the $k$-th client's confidence score on sample $\mathbf{x}$. 
Note that in practical applications, local models $\{f_k\}_{k=1}^K$ are typically developed using their local resources and proprietary techniques, thereby incurring the challenges related to model heterogeneity, data privacy,  
and the model intellectual property concerns.

\subsection{Federated Voting}

FedAvg is a notorious federated learning framework that collects the model parameters from clients and aggregates them by averaging. However, FedAvg assumes that all local models have the same structure and can easily lead to privacy leakage through the shared local model parameters.

\begin{figure}\centering
    \includegraphics[scale=0.25]{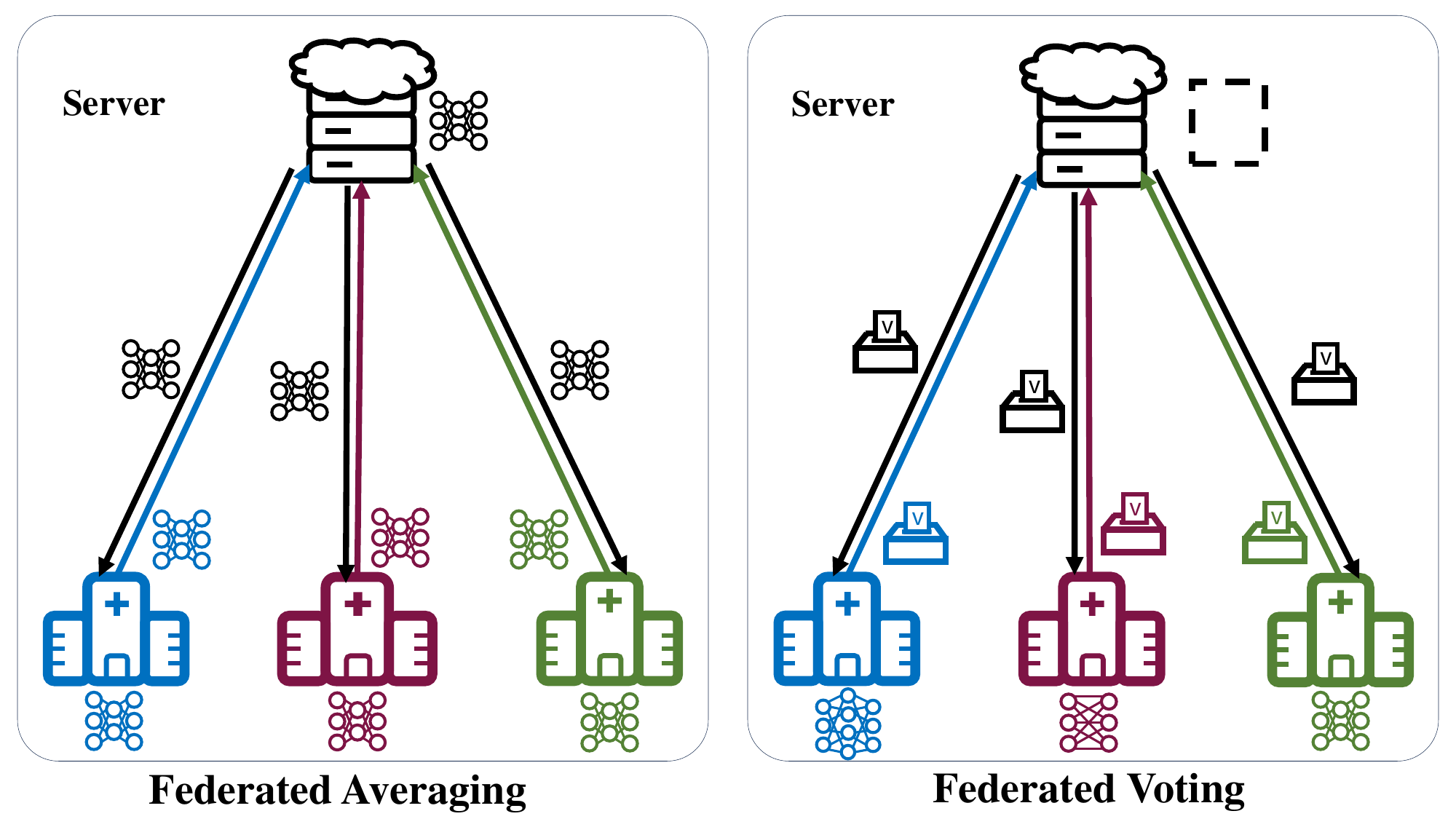}
    \caption{A comparison between FedAvg and the federated voting.}
    \label{fig:Fed_comparison}
\end{figure}

In order to  address the model heterogeneity issue and the data privacy concern simultaneously, we introduce a federated voting framework that allows local clients to share their votes (instead of model parameters) on an auxiliary unlabeled public dataset for aggregation. 
A comparison between this federated voting framework and FedAvg is illustrated in Figure \ref{fig:Fed_comparison}.

This work will exploit a novel abstention-aware federated voting mechanism, which collects high-confidence local votes to generate global votes (pseudo labels) for an auxiliary unlabeled public dataset for collaboration. Specifically, each client first uses its local private model to make predictions on unlabeled dataset and then votes using a novel threshold-based abstention-aware voting mechanism (see Section \ref{sec:method}). In particular, the voting mechanism only collects high-confidence predictions, in the sense that a low-confidence prediction score leads to an abstention. The central server collects these local votes and then takes the majority to generate a global vote, which will be regarded as pseudo labels for the unlabeled dataset. The consolidated global votes are then distributed to the clients to proceed on the collaborative learning.

It is worth noting that the proposed federated voting framework resolves the issue of model heterogeneity and does not require sharing local model parameters, thereby protecting the intellectual property of local models as well. Furthermore, the voting mechanism does not share the exact local predictions, which protects data privacy to a certain degree. By employing a local differential privacy mechanism as shown in the following subsection, data privacy protection could be further strengthened.

\subsection{Local Differential Privacy}
Local differential privacy is a mathematical approach for protecting the privacy of individual clients in a federated learning framework, see e.g. \cite{dpepsilon}. In many practical scenarios, the central server is an untrusted third party.
Local differential privacy refers to that each client perturbs his/her local data records to satisfy differential privacy, and sends only the randomized, differentially private version of the data records to the central server for aggregation.

  A randomized algorithm $\mathcal{A}_k$ for the $k$-th client satisfies $\epsilon$-\textit{local differential privacy} (LDP) if and only if for any two inputs $t$ and $t'$ in the domain of $\mathcal{A}_k$, and for any output $\Tilde{t}$ of $\mathcal{A}_k$, we have
\begin{equation*}
    \text{Pr}[\mathcal{A}_k(t) =\Tilde{t}\, ] \leq e^{\epsilon} \cdot \text{Pr}[\mathcal{A}_k(t') =\Tilde{t}\,]
\end{equation*}
where $\epsilon$ is a non-negative real number, often called the \textit{privacy budget} representing the strength/level of privacy guarantee. The smaller the privacy budget $\epsilon$ is, the more stringent the privacy guarantee becomes.

The piecewise mechanism is a simple,  high-performance and less-biased mechanism achieving LDP by virtue of an $\epsilon$-based piecewise function \cite{pmdp}. 
Given a real number $t\in[-1, 1]$, the piecewise mechanism outputs a perturbed value $\Tilde{t}\in [-T, T]$, where 
\begin{align}
    T \triangleq \frac{e^{\epsilon/2} + 1}{e^{\epsilon/2} - 1}.
\end{align}
The input $t$ is perturbed to $\Tilde{t}$ with a piecewise constant probability density function (pdf) as 
\begin{equation}
    \text{pdf}(\Tilde{t}=x|t) = \begin{cases}
       \rho,& \text{if}\ x \in [l(t),r(t)], \\
       \frac{\rho}{e^{\epsilon}},& \text{if}\ x\in [-T,l(t))\cup (r(t), T]\\
    \end{cases}
\end{equation}
where 
\begin{align}
    \rho &= \frac{e^{\epsilon}-e^{\epsilon/2}}{2 e^{\epsilon/2}+2}, \\
    l(t) &= \frac{T+1}{2} \cdot t - \frac{T-1}{2}, \\
    r(t) &= l(t) + T-1.
\end{align}

The piecewise mechanism 
is illustrated in Algorithm \ref{alg:Piecewise}, which will be used as a building block in Section \ref{sec:method}. 

\begin{algorithm}
\caption{Piecewise Mechanism  \cite{pmdp}.}
\label{alg:Piecewise}
\textbf{Input:} {Real number $t\in [-1, 1]$, privacy budget $\epsilon$} \\
\textbf{Output:} {Perturbed real number $\Tilde{t}$} \\
1. Uniformly sample $\alpha$ from $[0,1]$ \\
2. \textbf{if} $\alpha < \frac{e^{\epsilon/2}}{e^{\epsilon/2}+1}$ \textbf{do}\\
3.   \quad Uniformly sample $\Tilde{t}$ from $[l(t),r(t)]$\\
4.  \textbf{else}\\
5.   \quad  Uniformly sample $\Tilde{t}$ from $[-T,l(t))\cup (r(t), T]$\\
6.   \textbf{return:} $\Tilde{t}$
\end{algorithm}

\subsection{Healthcare Datasets}
\label{subsec:datasets}
Disease diagnosis and predicting patient risk of their health condition are crucial for tailoring appropriate care in hospital \cite{FGLB2018}. 
In this paper, we leverage two real-world healthcare datasets to evaluate the proposed framework: the diabetes dataset \cite{diabetes} and the Medical Information Mart for Intensive Care III (MIMIC-III) dataset \cite{mimic3}.

\textit{1)} 
For diabetes prediction, we will utilize a popular and simple diabetes dataset from the National Institute of Diabetes and Digestive and Kidney Diseases. This dataset contains 768 instances of female patients, all over 21 years old and of Pima Indian heritage. 
The primary objective is to predict whether a patient has diabetes based on various diagnostic measurements. 
The diagnoses are categorized as 0 (negative to diabetes) or 1 (positive to diabetes). 
Despite its limited number of features, this dataset offers a straightforward and insightful basis for evaluating the learning framework in healthcare.

\textit{2)}
For mortality prediction in the intensive care unit (ICU), we will utilize the MIMIC-III dataset, managed by the MIT Laboratory for Computational Physiology. 
This publicly accessible database contains comprehensive electronic health records (EHR) covering over 53,000 hospital admissions at Beth Israel Deaconess Medical Center from 2001 to 2012. 
It encompasses detailed records of approximately 40,000 unique patients aged 16 and above, including 4,579 charted observations and 380 laboratory tests. 
This dataset provides extensive time-series data for each patient, documenting a wide array of medical interactions such as procedures, medications, and diagnoses, along with more intricate data like medical notes. 
Given its comprehensive nature and larger volume, MIMIC-III indicates a more complex task  compared to the diabetes dataset. Our experiments aim to predict in-hospital patient mortality within the first 24 hours of admission, labeling the outcome as 0 for survival and 1 for mortality.

\section{Methods}
\label{sec:method}

\subsection{The Proposed AAFV}
To address the model heterogeneity, the model intellectual property and the data privacy challenges, we introduce an \textit{Abstention-Aware Federated Voting (AAFV)} framework, which integrates the piecewise mechanism 
and a novel abstention-aware voting mechanism 
to simultaneously guarantee the privacy and confidentiality of local clients. 
The overall AAFV framework is illustrated in Algorithm \ref{alg:FLprogram}. 
More precisely, we describe it from the following specific phases.

\vskip 0.02cm 
\textbf{Pre-train:} 
Since federated learning is a collaborative system, free-riders are not tolerated.
Therefore, prior to the communication for collaborative learning, each client pre-trains their local model $f_k$ on device using a private labeled dataset $\mathcal{D}_k$. 
This process is conducted as the typical supervised learning 
by aligning the prediction for each sample $\mathbf{x}_i\in \mathbb{R}^C$ with its corresponding ground-truth label $y_i \in \{0, 1\}$. 

\textbf{Local vote:} 
An auxiliary public unlabeled dataset $\mathcal{D}^u\subseteq \mathbb{R}^C$
will be employed to bridge the heterogeneous local models in collaborative learning.
First, 
each client $k$ generates their predictions $\mathbf{p}_{k}=(\mathbf{p}_{k,1},\ldots,\mathbf{p}_{k,N_u}) \in [0,1]^{N_u}$ 
for the unlabeled dataset $\mathcal{D}^u$, where $\mathbf{p}_{k,i}$ represents the confidence score from the $k$-th client for the $i$-th sample being classified as positive. 
Next, 
to enhance the privacy protection, 
the local predictions $\mathbf{p}_{k}$ are perturbed to $\Tilde{\mathbf{p}}_k=(\Tilde{\mathbf{p}}_{k,1},\ldots,\Tilde{\mathbf{p}}_{k,N_u}) \in\mathbb{R}^{N_u}$ using the piecewise mechanism shown in Algorithm \ref{alg:Piecewise}. 

Then 
each client casts their local vote $\mathbf{v}_k=(\mathbf{v}_{k,1},\ldots,$ $\mathbf{v}_{k,N_u})\in \{0,1,*\}^{N_u}$ based on $\Tilde{\mathbf{p}}_k$. 
To mitigate the effects of perturbation and to select high-confidence predictions from local votes, we introduce an abstention-aware voting mechanism described as
\begin{equation}\label{eq:pl}
    \mathbf{v}_{k,i} = \begin{cases}
    0, &   \Tilde{\mathbf{p}}_{k,i} \leq \tau \\
    *, & \tau < \Tilde{\mathbf{p}}_{k,i} < 1-\tau  \\
    1, &   \Tilde{\mathbf{p}}_{k,i} \geq 1-\tau
    \end{cases} 
\end{equation}
where $\tau\in (0,1)$ is a predetermined threshold to identify valid high-confidence predictions and 
the value of $\mathbf{v}_{k,i}$ is the vote from client $k$ for the $i$-th sample. 
If 
$\mathbf{v}_{k,i} = 1$ (resp. $0$), 
the client $k$ asserts that the $i$-th sample is positive (resp. negative)
with high confidence. 
If $\mathbf{v}_{k,i} = *$, 
it indicates that client $k$ does not have enough confidence in the prediction for sample $i$, and hence makes an abstention vote to avoid introducing confusion into the collaborative learning process.
Building on this, the local clients upload their votes to the central server for further aggregation. 

\textbf{
Consolidate:} 
The central server collects all the uploaded local votes for the unlabeled dataset and then proceeds to conduct consolidation process. To that end, the server first counts the total number of positive votes from all clients for each sample $i$, denoted as $\sum_{k=1}^{K} \mathbb{I}(\mathbf{v}_{k,i}=1)$, where $\mathbb{I}(\mathbf{v}_{k,i}=1)$ is an indicator function that equals $1$ if $\mathbf{v}_{k,i}=1$ is true and equals $0$ otherwise. 
Similarly, the server also counts the negative votes in the same manner. 
All the local abstention votes are regarded invalid and excluded from this count.

Then the central server generates consolidated votes $\Bar{\mathbf{v}}=(\Bar{\mathbf{v}}_1,\ldots,\Bar{\mathbf{v}}_{N_u})\in\{0,1,*\}^{N_u}$ by determining the majority from the local votes as the global votes for the unlabeled dataset. This is achieved by
\begin{equation}\label{eq:globalvote}
    \Bar{\mathbf{v}}_{i} = \begin{cases}
    0, &  \sum_{k=1}^{K} \mathbb{I}(\mathbf{v}_{k,i}=0) > \sum_{k=1}^{K} \mathbb{I}(\mathbf{v}_{k,i}=1)\\
    1, & \sum_{k=1}^{K} \mathbb{I}(\mathbf{v}_{k,i}=1) > \sum_{k=1}^{K} \mathbb{I}(\mathbf{v}_{k,i}=0) \\
    *, & \text{otherwise} \\
    \end{cases} 
\end{equation}
where $\Bar{\mathbf{v}}_{i} = *$ indicates an abstention global vote due to the non-separable local voting results. 
These consolidated votes, now serving as pseudo labels for the unlabeled data, will be sent back to local clients for further learning.

\textbf{Revisit:} 
Each local client receives consolidated pseudo labels for the unlabeled data from the central server. 
By removing all invalid samples marked with a pseudo label $*$ from $\mathcal{D}^u$, a high-confidence  pseudo labeled dataset $\hat{\mathcal{D}}^u$ can be created. 
Accordingly, each client utilizes the combination of $\hat{\mathcal{D}}^u$ and their own private dataset to enhance their local model. 
Again, the training process here is conducted as supervised learning by aligning the model predictions with their corresponding ground-truth or pseudo labels.

This methodology allows local clients to learn from each other without sharing their private data and the details of local models, thereby preserving the confidentiality and intellectual property of heterogeneous models involved.
The proposed AAFV framework is summarized in Algorithm \ref{alg:FLprogram}.

\subsection{Analysis of AAFV}
The proposed AAFV framework provides an efficient and confidential federated voting scheme that effectively addresses both model heterogeneity and data privacy challenges. More precisely, AAFV offers the following advantages.

\begin{itemize}
    \item Privacy. 
    AAFV does not require clients to share their private data, thereby preventing direct privacy leakage. 
    Moreover, AAFV adheres to local differential privacy principles by implementing the piecewise mechanism, which 
    further strengthens the privacy protection of private data 
    against membership inference attacks. 
\end{itemize}

\begin{algorithm}
\caption{The proposed AAFV framework}\label{alg:FLprogram}
\textbf{Input:} Public unlabeled dataset $\mathcal{D}^u$, local labeled dataset $\mathcal{D}_k$, heterogeneous local model $f_k, (k=1, $ $ \dots, K)$, communication epoch number $E_{com}$.\\
\textbf{Output:} Heterogeneous local models $f_k, (k=1,$ $\dots, K)$\\
1. \textbf{Pre-train:} Each client pre-trains local model $f_k$ with private labeled dataset $\mathcal{D}_k$. \\
2. \textbf{for} $e=1,\dots, E_{com}$ \textbf{do}\\
3. \text{\quad \ }  \textbf{for} $k=1,\dots, K$ \textbf{do}\\
4. \text{\quad \ } \text{\quad \ } $\mathbf{p}_k \gets f_k(\mathcal{D}^u)$\\
5. \text{\quad \ } \text{\quad \ } $\Tilde{\mathbf{p}}_k \gets$ Piecewise Mechanism$(\mathbf{p}_k)$ via Alg. \ref{alg:Piecewise}\\
6. \text{\quad \ } \text{\quad \ } $\mathbf{v}_k \gets$ Abstention-Aware Voting$(\Tilde{\mathbf{p}}_k) $ via \eqref{eq:pl}\\
7. \textbf{Consolidate:} The central server counts valid votes from
clients and consolidates them to global votes $\Bar{\mathbf{v}}$ via \eqref{eq:globalvote}. \\
8.  \textbf{Revisit:} Each client $k$ utilizes the combination of the
pseudo labeled public dataset $\hat{\mathcal{D}}^u$
and their private  dataset $\mathcal{D}_k$ to enhance their local model $f_k$.\\
9. \textbf{return:} $f_k, (k=1,\dots, K)$
\end{algorithm}


\begin{itemize}
    \item Heterogeneity. AAFV employs a novel federated voting mechanism with an auxiliary unlabeled dataset for aggregation, which does not demand the structure consistency of local models, making it particularly effective in handling heterogeneous models in the collaborative learning.
    \item Confidentiality. 
    Throughout the training process of AAFV, the detailed architectures of local models remain undisclosed, thereby effectively protecting the confidentiality and intellectual property of these models. 
    \item Effectiveness. 
    The abstention-aware voting mechanism used in AAFV selects high-confidence predictions from local clients to improve the effectiveness of collaborative learning. Additionally, the piecewise mechanism introduces noise with smaller variance, affecting the model utility less than other LDP algorithms.
    The real-world performance of AAFV is referred to Section \ref{sec:experiment}.
    \item Low-cost. 
    In the learning process, clients only upload local votes and download global votes, which are all one-dimensional and much less expensive in communication compared with sharing model parameters. Hence, AAFV offers the benefit of low communication cost, in particular for large-scale models.  
\end{itemize}

\section{Experiments}
\label{sec:experiment}
\subsection{Diabetes Prediction}
Diabetes, a prevalent disease, significantly increases the risk of heart disease, eye problems, and other severe health issues. Machine learning offers a promising solution to the early diagnosis of diabetes by effectively analyzing hospital medical records and personal healthcare data. This early detection of diabetes not only enables potential patients to seek timely medical treatment but also encourages them to adopt necessary lifestyle adjustments \cite{mldiabetes}.
We will conduct experiments on diabetes prediction
with the following components.

\begin{figure}\centering
    \includegraphics[scale=0.33]{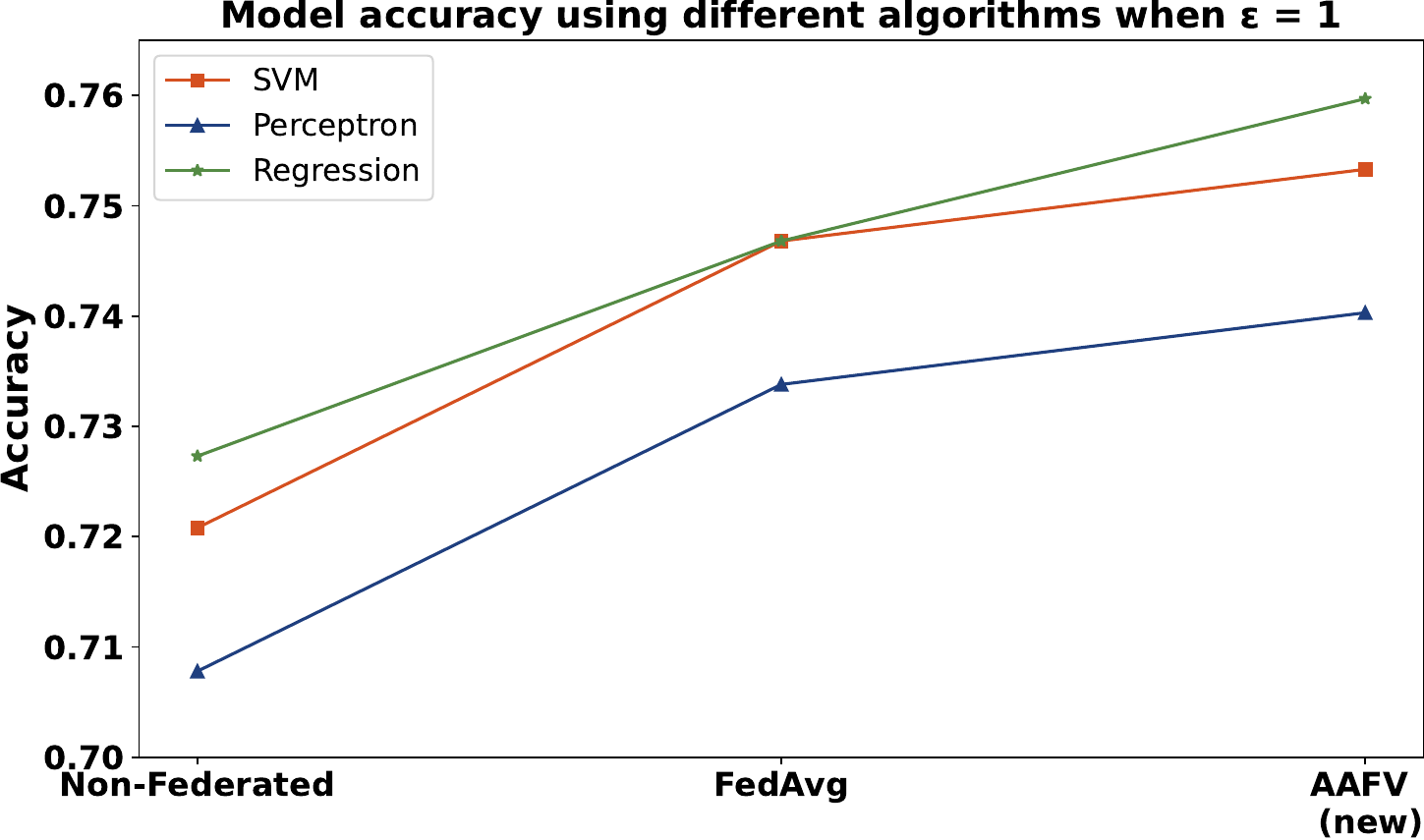}
    \caption{Accuracy on diabetes prediction across different scenarios when $\epsilon=1$.}
    \label{fig:diabetesacc}
\end{figure}

\subsubsection{Dataset}
As described in Section \ref{subsec:datasets}, the diabetes dataset comprises 768 samples \cite{diabetes}.
We randomly draw 153 samples from it as the test dataset, and draw another 126 samples as the unlabeled dataset. 
Then we divide the remaining 489 samples into three parts. Each contains 163 samples as the private labeled datasets.

Each sample consists of 8 features: number of times pregnant, plasma glucose concentration at two hours, diastolic blood pressure, triceps skin fold thickness, 2-hour serum insulin, BMI, diabetes pedigree function value, and ages.
To avoid numerical instability across the features, we apply a standard z-score normalization, by subtracting the mean from each feature and then dividing the feature values by their standard deviation. 
Consequently, each feature is transformed to have a mean of 0 and a standard deviation of 1. 

\subsubsection{Model Architectures}

In the diabetes prediction task, we set up 3 clients with three heterogeneous models according to the dataset scale. We use three typical machine learning models in the experiments: a support vector machine (SVM) \cite{svm}, a perceptron \cite{perceptron}, and a logistic regression \cite{regression}, for three different clients, respectively.

\subsubsection{Implementation} 
To achieve a comprehensive benchmark result, experiments are conducted across three scenarios: AAFV in a heterogeneous scenario, FedAvg in a homogeneous scenario, and individual models in a non-federated scenario. In the heterogeneous scenario, the AAFV framework incorporates three different models – a SVM, a perceptron, and a logistic regression – serving as local clients. For the homogeneous scenario, the FedAvg framework is implemented three times, each with identical model structures.
In the non-federated scenario, we train each of the three models independently using their respective local private labeled datasets.
In the experiments, each framework, AAFV or FedAvg, conducts $E_{com}=30$ communication epochs. 
In the non-federated scenario, each model is trained with $300$ epochs.

\subsubsection{Evaluation}
We evaluate the performance using the average test accuracy of 50 independent experiments with distinct random seeds to simulate the randomness. 
To facilitate a fair comparison, the privacy budget $\epsilon$ is uniformly set at $1.0$ for all experiments. In the homogeneous and non-federated scenarios, we implement a Laplacian mechanism to perturb the local model parameters to achieve differential privacy.

\subsubsection{Results}
Figure \ref{fig:diabetesacc} demonstrates the averaged test accuracy of SVM, perceptron, and logistic regression under three different scenarios. 
Notably, the proposed AAFV outperforms FedAvg and non-federated frameworks in all the three models.

\begin{figure}\centering
    \includegraphics[scale=0.48]{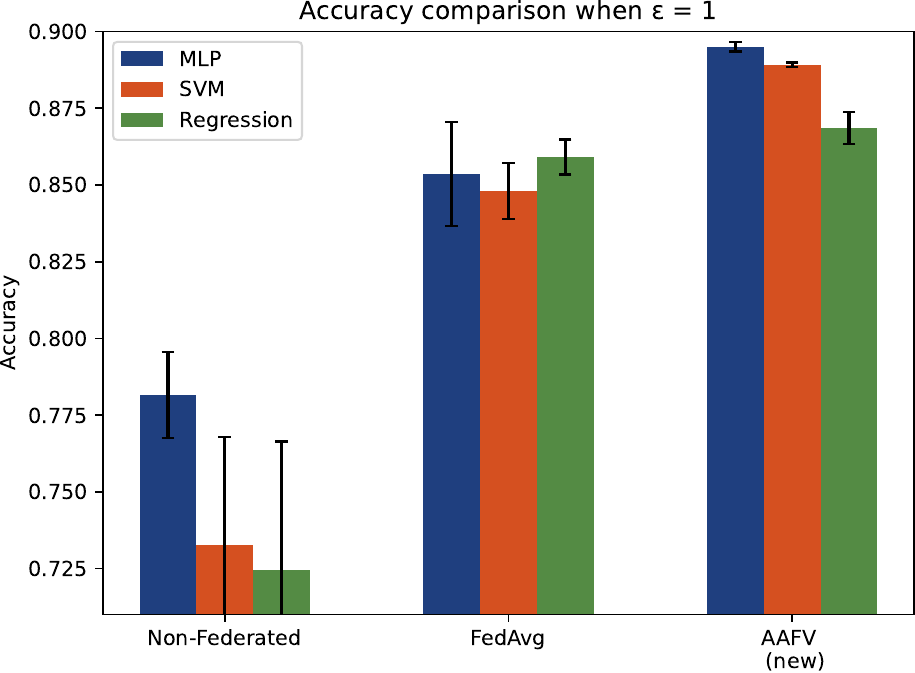}
    \caption{Accuracy on MIMIC-III across different scenarios when $\epsilon=1$.}
    \label{fig:mimicacc}
\end{figure}

\subsection{In-hospital Patient Mortality Prediction}
\subsubsection{Dataset}
EHRs offer researchers and healthcare facilities an opportunity to investigate the in-hospital patient information. By exploiting the massive EHRs dataset, hospitals are able to evaluate the effectiveness of current treatments and to improve patient healthcare strategies.
In this experiment, we extract the features of MIMIC-III dataset following \cite{MimicExtract}, and attain an in-hospital mortality dataset with $23,944$ samples which is composed of $7,488$ features. We divide the dataset into a test dataset with $4,789$ samples, and training dataset with $19,155$ samples. The training dataset is further divided into an unlabeled dataset with $3,831$ samples, and three local labeled datasets, each containing $5,108$ samples. 

\subsubsection{Model Architectures}
Due to the extensive feature space of MIMIC-III dataset, Multilayer Perceptron (MLP) \cite{mlp} is employed.
For the in-hospital patient mortality prediction task, the MLP configuration is a three-layer network comprising: an input layer matching the dimension of the feature space; a hidden layer, with a size that is half of the number of features and output dimension, and an output layer consisting of a single output for the binary classification task. Each layer is followed by an activation layer employing the Rectified Linear Unit (ReLU) function, which contributes to more efficient computation and better fitting result.
In addition to the MLP model, SVM and logistic regression are also utilized in
the experiments for a comparative analysis.

\subsubsection{Results}
As it is illustrated in Figure \ref{fig:mimicacc}, the AAFV framework demonstrates a stable, effective performance with a privacy budget of $\epsilon=1$. The accuracy across all three models demonstrates a notable improvement of $3\%$ on average. In particular, the MLP and SVM models within the AAFV framework significantly surpass the performance of those trained under the FedAvg framework with p-value averaged as $0.0025$, below the typical significance threshold of $0.05$. 

\section{Conclusion}
This paper explored the heterogeneous federated learning with local differential privacy, introducing a new framework termed AAFV.
This framework is designed to collaboratively and confidentially train local clients with heterogeneous structures. 
AAFV utilizes a novel abstention-aware voting mechanism with a threshold-based abstention method that selects high-confidence votes from local models. 
Experiments on two real-world healthcare datasets indicate that AAFV consistently outperforms the typical FedAvg framework and non-federated scenarios in test accuracy while preserving the same privacy level. 
It is of interest to further verify the performance of AAFV in more real-world healthcare applications. 

\section*{Acknowledgment}
This work was supported by WISE program (MEXT) at Kyushu University, and Toyota Riken Scholar program.


\end{document}